# An Algorithm for Finding Minimum d-Separating Sets in Belief Networks

**Silvia Acid**      **Luis M. de Campos**
Depto. de Ciencias de la Computación e I.A.
Universidad de Granada
18071-Granada, Spain

## Abstract

The criterion commonly used in directed acyclic graphs (dags) for testing graphical independence is the well-known d-separation criterion. It allows us to build graphical representations of dependency models (usually probabilistic dependency models) in the form of belief networks, which make easy interpretation and management of independence relationships possible, without reference to numerical parameters (conditional probabilities). In this paper, we study the following combinatorial problem: finding the minimum d-separating set for two nodes in a dag. This set would represent the minimum information (in the sense of minimum number of variables) necessary to prevent these two nodes from influencing each other. The solution to this basic problem and some of its extensions can be useful in several ways, as we shall see later. Our solution is based on a two-step process: first, we reduce the original problem to the simpler one of finding a minimum separating set in an undirected graph, and second, we develop an algorithm for solving it.

## 1 INTRODUCTION

Belief networks have become common knowledge representation tools capable of representing and handling independence relationships. The reasons for the success of belief networks are their ability to efficiently perform correct inferences by using local computations, and their reduced storage needs. Indeed, independence can modularize knowledge in such a way that we only need to consult the pieces of information relevant to the specific question in which we are interested, instead of having to explore a whole knowledge base; moreover, the storage requirements of, for example, a joint probability distribution are usually excessive, whereas the memory requirements of a suitable factorization of this distribution, taking into account the independence relationships, may be much smaller.

The capacity of belief networks for representing independence statements is based on the well-known graphical independence criterion called d-separation: if the dependency model satisfies certain properties then we can assert and use many (or all) of the independencies which are true in the model by checking for d-separation statements in the corresponding dag. However, d-separation is quite a subtle concept, more difficult to manage and interpret than its counterpart for undirected graphs, i.e., the separation criterion. In this paper we propose and solve an optimization problem related to d-separation. The basic problem may be formulated as follows: *given a pair of nodes, $x$ and $y$, in a dag, $G$, find the set of nodes with minimum size that d-separates $x$ and $y$*. An obvious extension of this basic problem is to find the minimum set d-separating two sets of nodes instead of two single nodes. A second extension of the basic problem is the following: given two sets of nodes, $X$ and $Y$, and given a third set of nodes, $Z$, find the minimum set, say $S$, that together with $Z$, d-separates $X$ and $Y$. It is worth noting that an algorithm for solving this last problem may also be used for testing d-separation: if the minimum set $S$ is empty, then $X$ and $Y$ are d-separated by $Z$; otherwise, they are not d-separated.

The first question we have to answer is: apart from the possible theoretical interest that this problem may have, does it have any practical utility? In our opinion the answer is positive. In general, the solution to this problem represents the minimum information which is necessary to know, in order to prevent two sets of nodes from influencing each other, either in the absence of any other information (first extension), or in the presence of some previous knowledge (second extension of the basic problem). Note that we are using the word *information* in a broad sense, not with the precise meaning attached to this term in Information Theory. We refer to the minimum number of (additional) variables, whose values we have to know in order to break the mutual influence between two sets of variables. More about this will be said in Section 5. The solution of our optimization problem can be useful in several ways:

Let us consider the following situation: we have a database containing instances of some variables in a given domain (or a joint probability distribution for these variables), and we also have a dag which is imagined to be an appropriate graphical representation for this domain. This means that the d-separation statements in the dag should correspond with true conditional independence statements in the domain. In order to test this assumption, we could select, for example, pairs of non adjacent nodes in the dag, next finding d-separating sets for these pairs, and then testing the corresponding conditional independencies in the database (or in the joint distribution). However, the complexity of testing these conditional independence statements increases exponentially (and, in the case of using a database, the reliability of the result decreases) according to the size of the d-separating sets. So, it may be quite interesting to select d-separating sets of a size as small as possible, instead of using some more obvious sets (such as, for example, the parent set of one of the two nodes in the pair).

Another possible application involves, in general, finding, in an already constructed network, the minimum number of variables preventing two given variables from influencing each other. We can outline several cases where this might be interesting: first, suppose that we are interested in obtaining information about a given variable (e.g., a classification variable), and we have to decide what other variables we should know in order to improve our information about the variable of interest; if knowing the value of each variable has a different cost, then a minimum (and inexpensive) set d-separating the variable of interest from other expensive variables would avoid the observation of these variables, by replacing it by the d-separating set. Second, given two variables representing, for example, two related diseases, we might be interested in finding the minimum number of variables (e.g., symptoms) that separate them. Third, if we are interested in a specific variable, and we know that the evidence transmitted by another variable is incorrect, then a minimum d-separating set might be the cheapest way of avoiding the propagation of the error to the variable of interest, without the need for propagating all the evidence accumulated again.

From these applications, the second one is rather vague, but we feel that an algorithm for solving our basic problem and its extensions may be a useful auxiliary tool for managing (large) belief networks. On the other hand, the first application addressed (i.e., given a graph, check whether it represents only independencies of a database or a distribution) has a more definite usefulness: an algorithm for learning belief networks from databases, based on the idea of finding minimum d-separating sets, has recently been designed (Acid-Campos 1996a). Let us briefly describe how minimum d-separating sets are used in the learning process: the basic idea is to measure (using cross-entropy) the discrepancies between the conditional independencies represented (through d-separation) in any given candidate network and the ones displayed by the database. The smaller these discrepancies are, the better the network fits the data. A measure of global discrepancy is then used as a heuristics by the search procedure, which explores the feasible solution space. The problem is that the number of d-separation statements displayed by a graph may be very high, and for efficiency and reliability reasons we want to exclude most of them, and use only a 'representative' set. The calculation of the minimum d-separating sets becomes important at this point: by using d-separating sets of minimum size we can reduce the computing time and also increase the reliability of the results. Depending on the network topology, the savings may be quite remarkable, even cutting down the exponential complexity in some cases (see (Acid-Campos 1996a) for details about the algorithm). For example, for the network displayed in Figure 1, if we were to test the conditional independence between any pair of nodes given the parent set of the largest numbered node in the pair, this would represent 10 tests of order zero, 30 tests of order one and 5 tests of order five. However, if we use the minimum d-separating set instead of the parent set, we need 40 tests of order zero and 5 tests of order one.

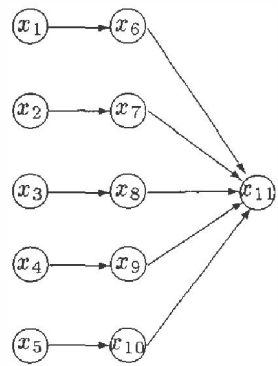

Figure 1: A dag with 11 nodes

The rest of the paper is organized as follows: in Section 2, we briefly describe several concepts which are basic for subsequent development, such as the separa-tion and d-separation criteria, as well as the relationship between them proposed in (Lauritzen et al. 1990). Section 3 shows how to reduce our original optimization problem about d-separation in dags to a simpler equivalent problem involving separation in undirected graphs. In Section 4, we propose an algorithm that, taking into account the previous results, finds a minimum d-separating set for any two non adjacent nodes in a dag. We also analize how to cope with the two proposed extensions of the basic problem. Finally, Section 5 contains the concluding remarks.

## 2  PRELIMINARIES

In this Section, we are going to describe the notation and some basic concepts used throughout the paper.

A *Dependency Model* (Pearl 1988) is a pair $M = (\mathcal{U}, I)$, where $\mathcal{U}$ is a finite set of elements or variables, and $I(.,.|.)$ is a rule that assigns truth values to a three place predicate whose arguments are disjoint subsets of $\mathcal{U}$. Single elements of $\mathcal{U}$ will be denoted by standard or Greek lowercase letters, such as $x, y, z, \alpha, \beta \ldots$, whereas subsets of $\mathcal{U}$ will be represented by capital letters, such as $X, Y, Z \ldots$ The intended interpretation of $I(X, Y|Z)$ (read $X$ is independent of $Y$ given $Z$) is that having observed $Z$, no additional information about $X$ could be obtained by also observing $Y$. For example, in a probabilistic model (Dawid 1979, Lauritzen al. 1990), $I(X, Y|Z)$ holds if and only if

$$P(\mathbf{x}|\mathbf{z}, \mathbf{y}) = P(\mathbf{x}|\mathbf{z}) \text{ whenever } P(\mathbf{z}, \mathbf{y}) > 0,$$

for every instantiation $\mathbf{x}$, $\mathbf{y}$ and $\mathbf{z}$ of the sets of variables $X$, $Y$ and $Z$.

On the other hand, a graphical representation of a dependency model $M = (\mathcal{U}, I)$ is a graph, $G = (\mathcal{U}, \mathcal{E})$, where $\mathcal{E}$ is the set of arcs or edges of $G$, such that the topology of $G$ reflects some properties of $I$. The way we relate independence assertions in $I$ with some topological property of a graph depends on the kind of graph we use; this property is *separation* for the case of undirected graphs (Lauritzen 1982, Pearl 1988) and *d-separation* for directed acyclic ones (Pearl 1988, Verma-Pearl 1990):

- *Separation*: Given an undirected graph $G$, two subsets of nodes, $X$ and $Y$, are said to be separated by the set of nodes $Z$, and this is denoted by $\langle X, Y|Z \rangle^s_G$, if $Z$ intercepts all the chains between the nodes in $X$ and those in $Y$, or, in other words, if the removal of the set of nodes $Z$ from the graph together with their associated edges, disconnects the nodes in $X$ from those in $Y$.

- *d-separation*: Given a dag $G$, a chain $C$ (a chain in a directed graph is a sequence of adjacent nodes, the direction of the arrows does not matter) from node $\alpha$ to node $\beta$ is said to be blocked by the set of nodes $Z$, if there is a vertex $\gamma \in C$ such that, either
  - $\gamma \in Z$ and the arrows of $C$ do not meet head to head at $\gamma$, or
  - $\gamma \notin Z$, nor has $\gamma$ any descendants in $Z$, and the arrows of $C$ do meet head to head at $\gamma$.

  A chain that is not blocked by $Z$ is said to be active. Two subsets of nodes, $X$ and $Y$, are said to be d-separated by $Z$, and this is denoted by $\langle X, Y|Z \rangle^d_G$, if all the chains between the nodes in $X$ and the nodes in $Y$ are blocked by $Z$.

In (Lauritzen al. 1990), an equivalent criterion to d-separation was proposed, which will be especially useful for our purposes. Several previous concepts are necessary to establish this equivalence: Let $G$ be a dag; given a node $\alpha$, the nodes $\beta$ such that there is a path in $G$ from $\alpha$ to $\beta$ are the *descendants* of $\alpha$, written $de(\alpha)$. Similarly, the nodes $\beta$ such that there is a path in $G$ from $\beta$ to $\alpha$ are called the *ancestors of* $\alpha$, denoted by $an(\alpha)$. A subset of nodes $X$ is an *ancestral set* if it contains its own ancestors, i.e., if $an(\alpha) \subseteq X \; \forall \alpha \in X$. We denote $\text{An}(X)$ as the smallest ancestral set containing $X$, i.e., $\text{An}(X) = X \cup (\cup_{\alpha \in X} an(\alpha))$. The *moral graph* $G^m$ of the dag $G$ is the undirected graph with the same set of nodes but with $\alpha$ and $\beta$ adjacent in $G^m$ if and only if they are adjacent in $G$ or if there is a node $\gamma$ such that $\alpha \to \gamma$ and $\beta \to \gamma$ are arcs in $G$. In other words, the moral graph is obtained from the original dag by 'marrying' parents with a common child and then dropping directions on arrows. Given a subset $S$ of nodes in a graph $G$, we denote $G_S$ to the subgraph of $G$ induced by $S$, i.e., a graph having only one subset $S$ of the set of nodes in the original graph but containing all the edges whose terminal nodes are both within this subset.

The following important result was proven in (Lauritzen al. 1990): Let $X$, $Y$ and $Z$ be disjoint subsets of nodes in a dag $G$. Then, $Z$ d-separates $X$ from $Y$ if and only if $Z$ separates $X$ from $Y$ in $(G_{\text{An}(X \cup Y \cup Z)})^m$, where $(G_{\text{An}(X \cup Y \cup Z)})^m$ is the moral graph of the subgraph of $G$ induced by the smallest ancestral set containing $X \cup Y \cup Z$. In symbols:

$$\langle X, Y|Z \rangle^d_G \iff \langle X, Y|Z \rangle^s_{(G_{\text{An}(X \cup Y \cup Z)})^m} \quad (1)$$

## 3  REDUCING d-SEPARATION TO SEPARATION

The problem we are trying to solve is the following: Find a d-separating set of minimum size for two sets of nodes $X$ and $Y$ in a given dag $G$.

This is a combinatorial optimization problem, and, in principle, it does not seem easy to solve, mainly because the d-separation criterion is difficult to manage, and is rather subtle: in some cases, the instantiation of some nodes (non head to head nodes) blocks the chains, and in others (head to head nodes) it unblocks them. So, we would like to transform the problem into an equivalent one, but avoiding the use of the d-separation criterion, and replacing it by a more 'uniform' criterion. The separation criterion for undirected graphs represents a good option. Therefore, the methodology we are going to use to solve our problem will first transform it into an equivalent separation problem.

The result in eq. (1) seems quite relevant to our purposes, because it relates d-separation with separation. However, this result is not directly applicable, because the transformed (undirected) graph where we have to test for the separation of $X$ from $Y$ depends on $X$, $Y$ and also on the d-separating set $Z$, the set we are looking for. Thus, we would have to test for the separation of $X$ from $Y$ given $Z$ in all the undirected graphs that can be formed by varying $Z$, and then selecting that $Z$ with minimum cardinality, a completely prohibitive brute force approach.

However, in this Section we prove that it is possible to transform our problem into a separation problem, where the undirected graph in which we have to look for the minimum set separating $X$ from $Y$ depends only on $X$ and $Y$. Later, in the next Section, we shall apply this result to developing an efficient algorithm that solves our original problem.

The next proposition shows that if we want to test a d-separation relationship between two sets of nodes $X$ and $Y$ in a dag, where the d-separating set is included in the smallest ancestral set of $X \cup Y$, then we can test this relationship in a smaller dag, whose set of nodes is formed only by the ancestors of $X$ and $Y$.

**Proposition 1** *Given a dag $G = (\mathcal{U}, \mathcal{E})$, $X, Y \subseteq \mathcal{U}$, and $Z \subseteq An(X \cup Y)$, let $H = G_{An(X \cup Y)}$ be the subgraph of $G$ induced by $An(X \cup Y)$. Then*

$$\langle X, Y | Z \rangle_G^d \iff \langle X, Y | Z \rangle_H^d$$

**Proof:** The necessary condition is obvious, because $H$ is a subgraph of $G$. Let us prove the sufficient condition: suppose that $\langle X, Y | Z \rangle_H^d$ but $\neg \langle X, Y | Z \rangle_G^d$. Then, in $G$, there is at least one chain, $C$, linking one node, $x$, in $X$ and one node, $y$, in $Y$, such that for all $\gamma \in C$, if $\gamma$ is not a head to head node then $\gamma \notin Z$, and if $\gamma$ is head to head, then either $\gamma \in Z$ or $de(\gamma) \cap Z \neq \emptyset$.

If the chain $C$ were only formed by nodes from $An(X \cup Y)$, then $C$ would be a chain in $H$ not blocked by $Z$, hence $\neg \langle X, Y | Z \rangle_H^d$, thereby contradicting the hypothesis. Therefore, there are nodes in $C$ which do not belong to $An(X \cup Y)$. Let $\gamma_0$ be one of these nodes, i.e., $\gamma_0 \in C$, $\gamma_0 \notin An(X \cup Y)$. As $\gamma_0$ belongs to a chain linking $x$ and $y$, and $\gamma_0$ is not an ancestor of $x$ or $y$, then $\gamma_0$ has to be either a head to head node of $C$ or an ancestor of a head to head node of $C$. As all the head to head nodes of $C$ belong to $Z$ or are ancestors of nodes that belong to $Z$, and since $Z \subseteq An(X \cup Y)$, then in either case $\gamma_0$ also belongs to $An(X \cup Y)$, which is again a contradiction. Therefore we have $\langle X, Y | Z \rangle_G^d$. □

The following proposition establishes the basic result necessary to solve our optimization problem: it says that any d-separating set for the sets of nodes $X$ and $Y$, which does not contain 'superfluous' nodes, must be included in the smallest ancestral set $An(X \cup Y)$. In other words, all the minimal d-separating sets are formed exclusively by nodes which are ancestors of either $X$ or $Y$.

**Proposition 2** *Given a dag $G = (\mathcal{U}, \mathcal{E})$, $X, Y \subseteq \mathcal{U}$, let $Z \subseteq \mathcal{U}$ be a set of nodes such that $\langle X, Y | Z \rangle_G^d$ and $\neg \langle X, Y | Z' \rangle_G^d$, $\forall Z' \subset Z$. Then $Z \subseteq An(X \cup Y)$.*

**Proof:** Let us suppose that $Z \not\subseteq An(X \cup Y)$. Let us define $Z' = Z \cap An(X \cup Y) \subseteq Z$. Then, from the hypothesis we have $\neg \langle X, Y | Z' \rangle_G^d$. As $Z' \subseteq An(X \cup Y)$, it is obvious that $An(X \cup Y \cup Z') = An(X \cup Y)$. Therefore, using eq. (1), we obtain

$$\neg \langle X, Y | Z' \rangle_{(G_{An(X \cup Y \cup Z')})^m}^s \equiv \neg \langle X, Y | Z' \rangle_{(G_{An(X \cup Y)})^m}^s$$

So, $X$ and $Y$ are not separated by $Z'$ in $(G_{An(X \cup Y)})^m$, hence there is a chain $C$ between $X$ and $Y$ in $(G_{An(X \cup Y)})^m$ that bypasses $Z'$, i.e., the chain $C$ is formed from nodes in $An(X \cup Y)$ which are outside of $Z$.

On the other hand, since $An(X \cup Y) \subseteq An(X \cup Y \cup Z)$, then $(G_{An(X \cup Y)})^m \subseteq (G_{An(X \cup Y \cup Z)})^m$. Then, the previously found chain $C$ is also a chain in $(G_{An(X \cup Y \cup Z)})^m$ that bypasses $Z$, which means that $X$ and $Y$ are not separated by $Z$ in $(G_{An(X \cup Y \cup Z)})^m$, and, once again using eq. (1), we obtain that $X$ and $Y$ are not d-separated by $Z$ in $G$, in contradiction to the hypothesis. Therefore it has to be $Z \subseteq An(X \cup Y)$. □

The next proposition shows that, by combining the results in propositions 1 and 2, we can reduce our original problem to a simpler one, which involves a smaller graph.

**Proposition 3** *Let $G = (\mathcal{U}, \mathcal{E})$ be a dag, and $X, Y \subseteq \mathcal{U}$. Then the problem of finding a minimum d-separating set for $X$ and $Y$ in $G$ is equivalent to the* *problem of finding a minimum d-separating set for X and Y in the induced subdag $G_{An(X \cup Y)}$.*

**Proof:** Let $H = G_{An(X \cup Y)}$, and let us define the sets $S_G = \{Z \subseteq \mathcal{U} \mid \langle X, Y|Z \rangle_G^d\}$ and $S_H = \{Z \subseteq An(X \cup Y) \mid \langle X, Y|Z \rangle_H^d\}$. Then we have to prove that

$$|T| = \min_{Z \in S_G} |Z| \Leftrightarrow |T| = \min_{Z \in S_H} |Z|.$$

From proposition 1, we deduce $S_H \subseteq S_G$, and therefore $\min_{Z \in S_H} |Z| \geq \min_{Z \in S_G} |Z|$.

-Necessary condition: If $|T| = \min_{Z \in S_G} |Z|$, then $\forall T' \subset T$ we have $T' \notin S_G$, and from proposition 2 we obtain $T \subseteq An(X \cup Y)$, and now using proposition 1 we get $T \in S_H$. So, we have $|T| \geq \min_{Z \in S_H} |Z| \geq \min_{Z \in S_G} |Z| = |T|$, hence $|T| = \min_{Z \in S_H} |Z|$.

-Sufficient condition: If $|T| = \min_{Z \in S_H} |Z| > \min_{Z \in S_G} |Z| = |Z_0|$, then we have $\forall Z' \subset Z_0$, $Z' \notin S_G$, and therefore, once again using propositions 2 and 1, we get $Z_0 \in S_H$, so that $|Z_0| \geq \min_{Z \in S_H} |Z| = |T|$, which is a contradiction. Thus, $|T| = \min_{Z \in S_G} |Z|$. □

The only remaining task is to transform our problem into an equivalent problem involving separation instead of d-separation:

**Theorem 1** *The problem of finding a minimum d-separating set for X and Y in a dag G is equivalent to the problem of finding a minimum separating set for X and Y in the undirected graph $(G_{An(X \cup Y)})^m$.*

**Proof:** Using the same notation from proposition 3, let $H^m$ be the moral graph of $H = G_{An(X \cup Y)}$, and $S_H^m = \{Z \subseteq An(X \cup Y) \mid \langle X, Y|Z \rangle_{H^m}^s\}$.

Let $Z$ be any subset of $An(X \cup Y)$. Then taking into account the characteristics of ancestral sets, it is clear that $H_{An(X \cup Y \cup Z)} = H$. Then, by applying eq. (1) to the graph $H$, we have

$$\langle X, Y|Z \rangle_H^d \Leftrightarrow \langle X, Y|Z \rangle_{(H_{An(X \cup Y \cup Z)})^m}^s \equiv \langle X, Y|Z \rangle_{H^m}^s$$

Hence $S_H = S_H^m$. Now, using proposition 3, we obtain

$$|T| = \min_{Z \in S_G} |Z| \Leftrightarrow |T| = \min_{Z \in S_{H^m}} |Z|. \qquad □$$

Before finishing off this Section, let us see in an example the practical significance of the previous results. Let us consider the graph in Figure 2, where we have numbered the nodes in an order compatible with the graph structure (i.e., the parents of any node appear before their children in the order). Let us suppose that we select the pair of nodes $x_3$ and $x_{15}$, and we want to find a set d-separating them with minimum size. Of course, we know that any node $x$ is d-separated from all the other nodes by the set of the parents of $x$, the children of $x$ and the parents of $x$'s children. So, in our case, we know that the two sets $\{x_1, x_2, x_4, x_5, x_6\}$ and $\{x_8, x_9, x_{10}, x_{16}, x_{17}, x_{18}\}$ d-separate $x_3$ from $x_{15}$. We also know that any node $x$ is d-separated from all its non-descendants by the set of $x$'s parents. Therefore, as $x_3$ is not a descendant of $x_{15}$, we can be sure that $x_3$ and $x_{15}$ are d-separated by the set $\{x_8, x_9, x_{10}\}$. However, can we find a smaller set that still d-separates these two nodes? To answer this question we would have to examine every possible chain linking $x_3$ and $x_{15}$, facing the enormous amount of d-separating sets that we can obtain from the graph $G$, and next select the set of minimum size.

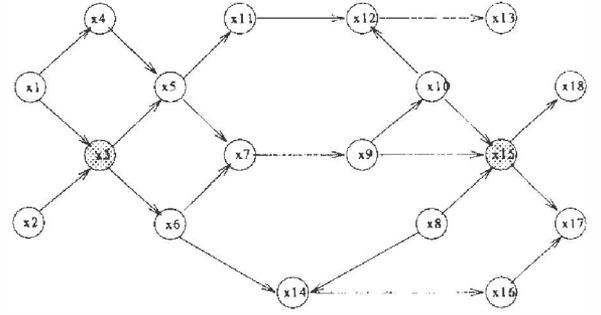

Figure 2: A dag $G$ with 18 nodes

The result in proposition 3 allows us to considerably reduce the search space where we have to look for the d-separating sets, by removing the nodes that do not belong to the set $An(x_3, x_{15})$. The subgraph $G_{An(x_3, x_{15})}$, which has 11 nodes, is shown in Figure 3. The complexity of the graph has decreased, thus reducing the number of chains we have to explore. Finally, the mor-alized graph $(G_{An(x_3, x_{15})})^m$, where, by virtue of theorem 1, we have to search for the minimum separating set of $x_3$ and $x_{15}$, is shown in Figure 4. From this graph, it is now apparent that the d-separating set of minimum size for $x_3$ and $x_{15}$ is $\{x_9\}$.

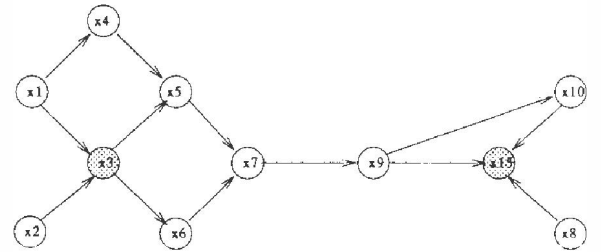

Figure 3: Dag $G_{An(x_3, x_{15})}$

## 4 THE ALGORITHM

In this Section, we develop an algorithm to solve the basic problem stated in the introduction. Later, we shall see how we modify this algorithm to solve the two proposed extensions of the basic problem.

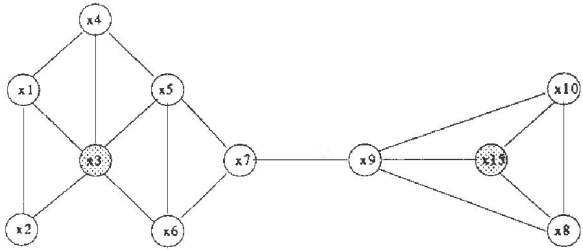

Figure 4: Moral graph $(G_{\mathrm{An}(x_3,x_{15})})^m$

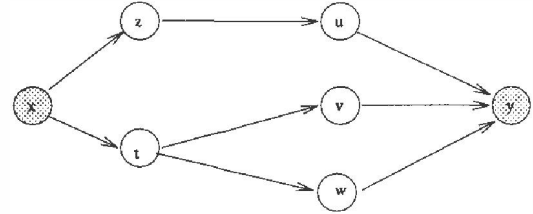

Figure 5: Dag where $x$ and $y$ can be d-separated by the sets $\{z,t\}$ and $\{u,t\}$

## 4.1 THE ALGORITHM FOR THE BASIC PROBLEM

We want to develop an algorithm for finding a minimum d-separating set for two given nodes $x$ and $y$ in a dag $G$. However, it may happen that there is more than one d-separating set for $x$ and $y$ with a minimum size. We could provide all these d-separating sets but, taking into account the kind of applications we have in mind, we prefer to find only one of them. In this case, we have to provide an additional criterion to select one minimum d-separating set. Our proposal is the following: as $G$ is an acyclic graph, either $x$ is not a descendant of $y$ or $y$ is not a descendant of $x$; suppose, for example, that $x$ is not a descendant of $y$. Then a natural d-separating set for $x$ and $y$ in $G$ would be the parent set of $y$, $pa(y)$; however, we want to select a set of minimum size. If $pa(y)$ had a minimum size, then this set would be chosen; otherwise, we should replace some of (or all) the parents of $y$ by other nodes, as long as this replacement diminishes the size of the d-separating set. But, in order to be coherent with our preference for the parent set of $y$, we should use nodes as near as possible to $y$ (in other words, a way of being sure that our algorithm will produce as its output the parent set of node $y$, should this set have a minimum size, is to design the algorithm with a built-in preference for nodes close to $y$). For example, for the dag displayed in Figure 5, to d-separate $x$ from $y$ using a set of minimun size, we would select the set $\{u,t\}$ instead of the set $\{z,t\}$. Therefore, our additional criterion for choosing only one minimum d-separating set is the following: *among all the minimum d-separating sets for $x$ and $y$, and supposing that $x$ is not a descendant of $y$, select the one which is nearer to $y$.*

Starting out from the results obtained in the previous Section, to solve the basic problem of finding a minimum d-separating set for a pair of nodes $x$ and $y$ in a dag $G$, it is suffice to find a minimum separating set for $x$ and $y$ in the undirected graph $(G_{\mathrm{An}(\{x,y\})})^m$.

So, bearing in mind the discussion above, our specific objective in this subsection will be the following: given an undirected graph $H = (\mathcal{V}, \mathcal{E})$, and given two nodes $x, y \in \mathcal{V}$, find a set of minimum size that separates $x$ from $y$; if there is more than one of these sets, select the one which is nearer to $y$ (the proximity being measured in terms of the length of the chains linking $y$ and the nodes in the separating set).

To design an algorithm for solving this problem we shall take advantage of the strong relationship that exists between problems of connectivity and flow problems in graphs. In general, a flow problem arises when, given a directed graph, we want to determine the value of the maximum flow that can be transmitted from a specified source node $s$ of the graph, to a specified sink node $t$. In this context, every arc of the graph has an associated number that represents the largest amount of flow that can be transmitted along the arc (the capacity of the arc). A method for solving this *maximum flow* problem was developed by Ford and Fulkerson (Ford-Fulker. 1979), which is based on an important result which establishes the relation between maximum flows and minimum cuts (Ford-Fulker. 1979): the value of the maximum flow from $s$ to $t$ in a graph is equal to the value of the minimum cut-set separating $s$ from $t$.

Unfortunately, the term cut-set does not refer to a separating set containing nodes but to a separating set containing arcs: a cut-set separating $s$ from $t$ is a set of arcs such that all the paths in the graph going from $s$ to $t$ must pass along some arc in the cut-set; the value of a cut-set is the sum of the capacities of its arcs. Therefore, if the capacity of each arc were 1, then the Ford-Fulkerson (F-F) algorithm would identify a cut-set of minimum size. So, although this result may be useful, it is not exactly the one we need: we are looking for separating node-sets in undirected graphs, whereas the previous result refers to separating arc-sets (cut-sets) in directed graphs.

However, there is a standard method for transforming separating arc-sets for directed graphs into separating node-sets for undirected graphs: we can see any undirected graph $H = (\mathcal{V}, \mathcal{E})$ as a directed one, $\vec{H} = (\mathcal{V}, \vec{\mathcal{E}})$, by simply considering every edge $u$–$v \in \mathcal{E}$ as the pair of arcs $u \to v$, $u \leftarrow v \in \vec{\mathcal{E}}$. Moreover, we can turn a problem of node connectivity in $\vec{H}$ into a problem of arc connectivity in an auxiliary graph $\vec{H}_{\text{aux}} = (\mathcal{V}', \vec{\mathcal{E}}_{\text{aux}})$, in the following way: (a) every node $u \in \mathcal{V}$ corresponds to two nodes $u^+, u^- \in \mathcal{V}'$; (b) for every arc $u \to v \in \vec{\mathcal{E}}$ corresponds an arc $u^- \to v^+ \in \vec{\mathcal{E}}_{\text{aux}}$; (c) we also introduce in $\vec{\mathcal{E}}_{\text{aux}}$ the arcs $u^+ \to u^-$. The transformation from graph $H$ into graph $\vec{H}_{\text{aux}}$ is illustrated in Figure 6. Moreover, we give all the arcs in $\vec{H}_{\text{aux}}$ a capacity equal to 1. Then, to search for a minimum node-set separating $s$ and $t$ in $H$ is equivalent to searching for a minimum cut-set separating $s^-$ from $t^+$ in $\vec{H}_{\text{aux}}$: since the total flow entering a node $u^+$ must, by necessity, travel along the arc $u^+ \to u^-$ whose capacity is 1, the maximum flow in the graph $\vec{H}_{\text{aux}}$ must correspond to a minimum cut-set containing only arcs of the form $u^+ \to u^-$; therefore, the nodes $u$ in $H$ that correspond to the arcs $u^+ \to u^-$ of the cut-set in $\vec{H}_{\text{aux}}$ comprise a separating node-set in $H$ of a size equal to the value of the cut-set.

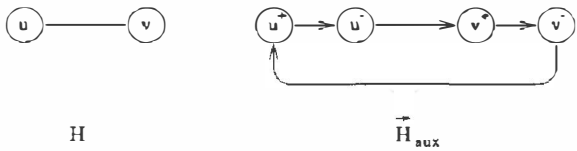

Figure 6: Transformation of $H$ into $\vec{H}_{\text{aux}}$

So, to solve the problem of finding a minimum set separating two nodes $x$ and $y$ in an undirected graph $H$, we can apply the F-F algorithm to find the maximum flow from $y^-$ to $x^+$ in the auxiliary directed graph $\vec{H}_{\text{aux}}$. The reason for using the maximum flow from $y^-$ to $x^+$ instead of the flow from $x^-$ to $y^+$ is that the very dynamics of the F-F algorithm will favor the presence of nodes close to the source $y^-$ in the separating set, as required. Fortunately, it is not really necessary to transform explicitly the original graph $H$ into the auxiliary graph $\vec{H}_{\text{aux}}$: we have developed an algorithm that works directly over $H$ (in fact, our implementation of the algorithm uses the original dag $G$ where we want to test for d-separation, and all the graph transformations, from $G$ to $H = (G_{\text{An}(X \cup Y)})^m$ and to $\vec{H}_{\text{aux}}$ are implicit). It is based on a suitable modification of the F-F algorithm. Due to space limitations, we shall not describe it here, see (Acid-Campos 1996b). Its complexity, like F-F's, is $O(nm^2)$, where $n$ is the number of nodes and $m$ the number of edges in the graph. It could be improved by using better maximum flow algorithms (McHugh 1990).

### 4.2 COPING WITH THE EXTENSIONS

In this subsection, we study how the algorithm developed above may be modified to deal with the extensions of the basic problem.

For the first extension, i.e., to find the minimum set d-separating two subsets of nodes $X$ and $Y$ (instead of two single nodes) in a dag $G = (\mathcal{U}, \mathcal{E})$, the solution is very simple: first, we build the undirected graph $(G_{\text{An}(X \cup Y)})^m = (\text{An}(X \cup Y), \mathcal{E}^m_{\text{An}(X \cup Y)})$, i.e., the moral graph of the subgraph of $G$ induced by the smallest ancestral set containing $X \cup Y$; next, starting out from this graph, we construct a new undirected graph $G^{XY} = (\mathcal{V}, \mathcal{F})$ as follows:

- $\mathcal{V} = \text{An}(X \cup Y) \cup \{\alpha_X, \beta_Y\}$,

- $\mathcal{F} = \mathcal{E}^m_{\text{An}(X \cup Y)} \cup \{\alpha_X\text{-}z \mid \exists x \in X \text{ s.t. } z\text{-}x \in \mathcal{E}^m_{\text{An}(X \cup Y)}\} \cup \{\beta_Y\text{-}y \mid \exists y \in Y \text{ s.t. } z\text{-}y \in \mathcal{E}^m_{\text{An}(X \cup Y)}\}$.

Put into words: we add two artificial (dummy) nodes $\alpha_X$ and $\beta_Y$, and connect $\alpha_X$ and $\beta_Y$ to those nodes that are adjacent to some node in $X$ and $Y$, respectively. It can be easily proven that

- $\langle \alpha_X, \beta_Y | Z \rangle^s_{G^{XY}} \iff \langle X, Y | Z \rangle^s_{(G_{\text{An}(X \cup Y)})^m}$, and

- if $\langle \alpha_X, \beta_Y | W \rangle^s_{G^{XY}}$ and $W \cap (X \cup Y) \neq \emptyset$, then $\langle \alpha_X, \beta_Y | W \setminus (X \cup Y) \rangle^s_{G^{XY}}$.

So, the separation of $X$ and $Y$ in $(G_{\text{An}(X \cup Y)})^m$ is equivalent to the separation of $\alpha_X$ and $\beta_Y$ in $G^{XY}$. Moreover, the minimum separating set for $\alpha_X$ and $\beta_Y$ in $G^{XY}$ cannot contain nodes from $(X \cup Y)$. Therefore, in order to find the minimum d-separating set for $X$ and $Y$ in $G$, it is suffice to find the minimum separating set for $\alpha_X$ and $\beta_Y$ in the auxiliary graph $G^{XY}$. So, we have reduced the problem to one of separation for single nodes, which can be solved using the previous algorithm.

The second extension of the basic problem was the following: given two sets of nodes $X$ and $Y$, and given a third set of nodes $Z$, find the minimum set, say $S$, such that $\langle X, Y | Z \cup S \rangle^d_G$. In this case, we try to find the minimum d-separating set for $X$ and $Y$ but with the restriction that some nodes in the d-separating set are fixed. It can be proven that all the propositions stated in Section 3 can be extended for dealing with this restriction, i.e., we can prove the following results (the proofs are almost identical to those in Section 3):

- If $S \subseteq \text{An}(X \cup Y \cup Z)$, and $H = G_{\text{An}(X \cup Y \cup Z)}$ then

$$\langle X, Y | Z \cup S \rangle^d_G \iff \langle X, Y | Z \cup S \rangle^d_H$$

- If $\langle X, Y | Z \cup S \rangle_G^d$ and $\neg \langle X, Y | Z \cup S' \rangle_G^d$, $\forall S' \subset S$, then $S \subseteq \text{An}(X \cup Y \cup Z)$.

So, by applying the result stated in (Lauritzen al. 1990) once again, the problem of finding, in a dag $G$, a minimum d-separating set for $X$ and $Y$, which contains the set $Z$, is equivalent to the problem of finding a minimum separating set for $X$ and $Y$, containing $Z$, in the undirected graph $(G_{\text{An}(X \cup Y \cup Z)})^m$. Now, it is suffice to eliminate the set $Z$ from this last graph, i.e., search for the minimum separating set for $X$ and $Y$ in the graph $((G_{\text{An}(X \cup Y \cup Z)})^m)_{\text{An}(X \cup Y \cup Z) \setminus Z}$.

## 5 CONCLUDING REMARKS

We have studied and solved the problem of finding minimum d-separating sets for pairs of variables in belief networks. Our method is based on a theoretical study that allows us to transform the original problem into an equivalent problem of separation in undirected graphs. The proposed algorithm implicitly uses this equivalence, and is based on a suitable modification of a well-known algorithm from the Operations Research literature, the Ford-Fulkerson algorithm of maximum flow in networks with capacities. We have also studied some extensions of the basic problem: finding minimum d-separating sets for subsets of variables, and finding minimum d-separating sets for variables or subsets of variables, with the restriction that some variables in the d-separating sets must be fixed. Our basic algorithm is also able to manage these extensions with minor modifications. Actual and potential applications of this research include learning belief networks from data (Acid-Campos 1996a) and problems related to the selection of the variables to be instantiated when using belief networks for inference tasks.

An interesting extension of this work would be to look for a d-separating set that has a minimum total state space instead of a minimum number of variables. Currently, we have ignored the domain size of each variable. However, it is clear that, in order to reduce the complexity of testing conditional independencies from data, it is more important to reduce the total state space than the number of variables involved in the d-separating set. From this perspective, a d-separating set of, for example, two binary variables would be preferred to one formed by a single variable having eight possible values. The methodology developed in this paper would also work for this problem: the only modification would be to change the capacity of a node to be the logarithm of its domain size instead of one (the algorithm developed in (Acid-Campos 1996b) should also be adapted accordingly).

From a more theoretical point of view, another interesting problem is the following: our algorithm can be seen as a method for finding, given a network, the conditional independence relationships involving the minimum number of variables. Can this process be reversed? In other words, given a Dependency Model (which may be or may be not dag-isomorphic), and given the set of all the conditional independence assertions between pairs of variables with conditioning sets of a minimum size, can we construct a graphical representation of this model, such that all the d-separation statements in the dag correspond with true conditional independence assertions in the model (i.e., an Independence map (Pearl 1988))? This would lead to the definition of a new concept, similar to that of *causal input list* or *recursive basis* (Verma-Pearl 1990), but with a more 'local' character. We plan to study these topics in the future.

### Acknowledgements

This work has been supported by the DGICYT under Project PB92-0939. We are grateful to two anonymous reviewers for their useful comments and suggestions.